\documentclass{article}

\usepackage{microtype}
\usepackage{graphicx}
\usepackage{subfigure}
\usepackage{booktabs} 
\usepackage{multirow}
\usepackage{float}
\usepackage{tcolorbox}

\usepackage{hyperref}

\usepackage[accepted]{icml2024}

\usepackage{amsmath}
\usepackage{amssymb}
\usepackage{mathtools}
\usepackage{amsthm}

\usepackage[capitalize,noabbrev]{cleveref}

\theoremstyle{plain}

\theoremstyle{definition}

\theoremstyle{remark}

\usepackage[textsize=tiny]{todonotes}

\icmltitlerunning{View From Above: A Framework for Evaluating Distribution Shifts in Model Behavior}

\begin{document}

\twocolumn[
\icmltitle{View From Above: \\
A Framework for Evaluating Distribution Shifts in Model Behavior}
\icmlsetsymbol{equal}{*}

\begin{icmlauthorlist}
\icmlauthor{Tanush Chopra}{equal,gt,apart}
\icmlauthor{Michael Li}{equal,cmu,apart}
\icmlauthor{Jacob Haimes}{apart}
\end{icmlauthorlist}

\icmlaffiliation{gt}{Georgia Institute of Technology}
\icmlaffiliation{cmu}{Carnegie Mellon University}
\icmlaffiliation{apart}{Apart Research}

\icmlcorrespondingauthor{Michael Li}{limichael353@gmail.com}
\icmlcorrespondingauthor{Tanush Chopra}{tanushchop@gmail.com}

\vskip 0.3in
]

\printAffiliationsAndNotice{\icmlEqualContribution}

\begin{abstract}
When large language models (LLMs) are asked to perform certain tasks, how can we be sure that their \textit{learned representations} align with reality? We propose a domain-agnostic framework for systematically evaluating \textit{distribution shifts} in LLMs decision-making processes, where they are given control of mechanisms governed by pre-defined rules. While individual LLM actions may appear consistent with expected behavior, across a large number of trials, statistically significant distribution shifts can emerge. To test this, we construct a well-defined environment with known outcome logic: blackjack. In more than 1,000 trials, we uncover statistically significant evidence suggesting behavioral misalignment in the learned representations of LLM.
\end{abstract}

\section{Introduction}

The field of machine learning has witnessed significant advancements in developing ``agentic'' systems that directly influence real-world outcomes \cite{durante2024agentaisurveyinghorizons}. Large language models (LLMs) have emerged as powerful tools in this domain, attracting considerable attention for their application to a wide range of tasks \cite{Wang_2024, dipalo2023unifiedagentfoundationmodels, vezhnevets2023generativeagentbasedmodelingactions, yao2023webshopscalablerealworldweb}. As large language models (LLMs) are increasingly deployed in real-world systems, there is growing concern about the potential for biased outcomes from these AI agents \cite{bender2021dangers, weidinger2021ethicalsocialrisksharm}. Much of the current research on LLM biases has focused on identifying whether these models exhibit the same types of biases found in humans---such as gender or racial biases---by assessing how they mimic patterns in their training data \cite{Caliskan_2017, abid2021persistentantimuslimbiaslarge}.

\begin{figure*}[h]
    \centering
    \includegraphics[width=\linewidth]{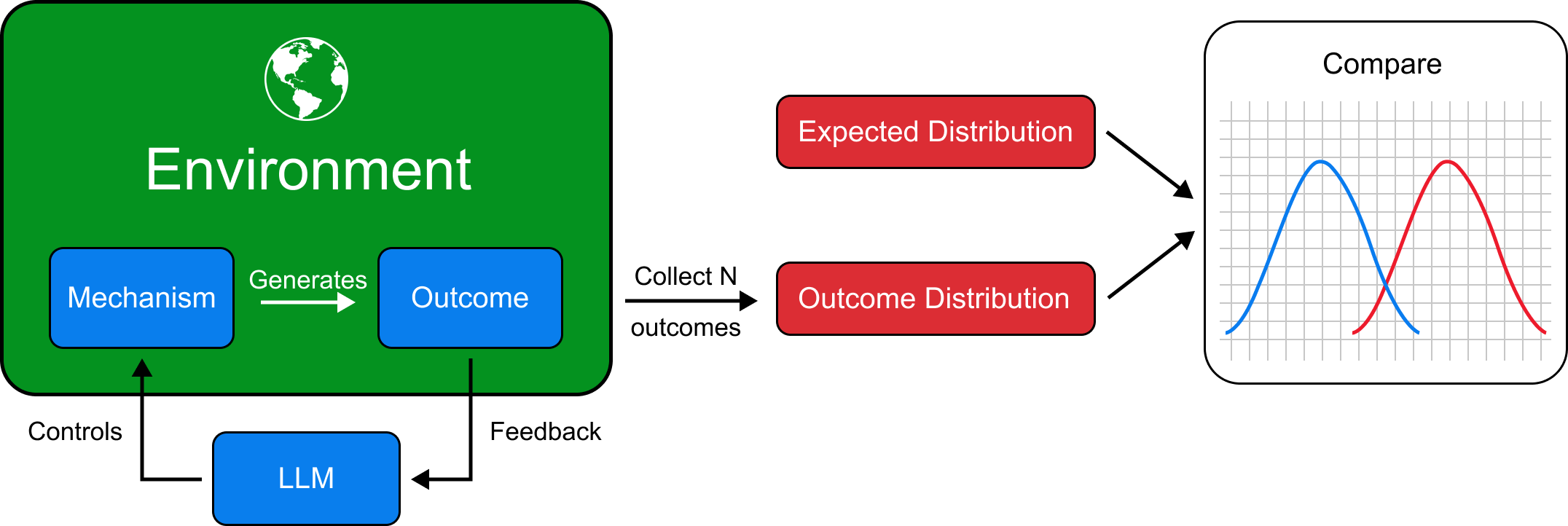}
    \caption{The VFA (View From Above) Framework for evaluating distribution shifts in LLM decision-making. We compare LLM-controlled outcomes in an environment against expected distributions. Statistical analysis is leveraged to detect potential LLM behavioral misalignment.}
    \label{fig:llm-evaluation-framework}
\end{figure*}

However, we suggest that this focus on human biases may overlook other dimensions of LLM behavior. While LLMs do mirror well-known human biases, it is important to consider that LLMs may also develop biases that differ from those of humans. There has been some previous work in this area, specifically comparing response bias in LLMs to humans \cite{tjuatja2024llms}. Understanding the nature of their biases---whether they align with human biases or diverge---is critical. To address this, we suggest a different approach for evaluating model behavior. 

We propose a domain-agnostic framework to detect \textbf{distributional} shifts in LLM decision-making within environments governed by known rules. By observing the LLM's control over a specific mechanism and comparing it to either theoretical or human generated outcomes, we establish a baseline for comparison. We then use statistical tests to compare the observed outcomes of the LLM, $P_{\text{observed}}(x_1, x_2, \ldots, x_n)$, to the expected outcomes, either theoretical or human generated, $P_{\text{expected}}(y_1, y_2, \ldots, y_n)$. This approach helps identify deviations in LLM behavior that may not be evident from individual actions.

As a proof-of-concept, we apply our framework to blackjack, drawing inspiration from methods used by casinos to detect anomalous behavior. Blackjack has straightforward rules and well-defined player behavior, allowing us to control all factors of the environment during our study. In this context, the mechanism we give the LLM control over is the drawing of cards from the deck. Through our investigation, we demonstrate that LLMs exhibit statistically significant deviations from expected outcomes in random card drawing, validating the effectiveness of our method.

While blackjack is a simplified example, it provides a clear picture of how our framework would be utilized. It is important to note that these initial results are not meant to reflect all LLM behavior, but rather illustrate the potential of our framework in detecting distribution shifts in LLM decision-making processes. Future work will extend the approach and make comparisons to human decision-making patterns.

\section{Related Works}

\textbf{LLMs and humans} Research into bias in LLMs has highlighted both the presence and potential causes of these biases \citep{bender2021dangers, blodgett-etal-2021-stereotyping, zou2023representationengineeringtopdownapproach}. Previous studies have primarily focused on identifying whether LLMs exhibit the same types of biases found in humans, such as gender or racial biases, by assessing how they mimic patterns in their training data \cite{Caliskan_2017, abid2021persistentantimuslimbiaslarge, lin2022truthfulqameasuringmodelsmimic}. However, LLMs may develop biases different from those of humans.

A study by \citet{lamparth2024humanvsmachinebehavioral} examined LLMs in high-stakes military decision-making through a wargame with 107 national security experts, revealing significant deviations between LLM-generated and human responses in a U.S.-China crisis scenario \cite{lamparth2024humanvsmachinebehavioral}. While LLMs and humans aligned at a high level, LLMs had different strategic tendencies and displayed more aggressive behaviors, especially in extreme scenarios. Our approach builds on this study's insights, improving quantitative comparison methods and extending applicability to a wider range of scenarios.

\textbf{Specific model evaluation methods.} Several works have contributed to the evaluation and understanding of LLM behavior in specific contexts, focusing on cognitive biases or inaccuracies introduced by prompt changes \cite{tjuatja2024llms, itzhak2023instructed}. While these approaches provide valuable insights, they do not directly compare LLM behavior against human behavior. Instead, they rely on previously collected human surveys to infer biases, rather than evaluating humans directly on the same datasets. Furthermore, by employing statistical tests to quantify distribution shifts, we are able to make more robust conclusions about differences in behavior.

\textbf{Behavioral science approach.} The concept of ``machine psychology'' has been introduced to study models using methods from behavioral science \cite{hagendorff2024machinepsychology}. Similar to our work, this approach involves a systematic evaluation involving LLM behavior. However, while machine psychology centers on directly evaluating LLM behavior, our methodology focuses on comparing this behavior to reality.

\section{Methods}

To test our hypothesis, we implemented a simplified version of blackjack and conducted a series of experiments comparing a control group with random card draws to an experimental group where card draws were controlled by a large language model (LLM). Figure \ref{fig:llm-evaluation-framework} illustrates our framework for evaluating distribution shifts in LLM decision-making.

\subsection{Blackjack Environment}

In our blackjack environment, a single player competes against a dealer using a standard 52-card deck, which is reshuffled before each hand. The player can choose between ``hit'' or ``stand'' actions. The dealer follows standard casino rules: hitting on 16 or below and standing on 17 or above. The game does not include advanced options such as splitting pairs, doubling down, or insurance bets. For a comprehensive overview of the rules, please refer to \Cref{appendix:blackjack_rules}.

\subsection{Experimental Setup}
We conducted two types of experiments: zero-shot and few-shot prompts, each evaluated with two different temperature settings.

Zero-shot experiments required the models to simulate card draws without prior examples, while the few-shot experiments provided the models with a small number of game examples before running 1,000 trials. Detailed prompting variations are provided in Appendix \ref{appendix:prompts}. Each experiment was performed at two temperature settings (temperature = 0, temperature = 0.5) to assess how model sampling influenced outcomes.

We tested several LLMs: \textbf{gpt-4o-2024-08-06}, \textbf{claude-3-5-sonnet-20240620} and \textbf{Llama 3 8B} \footnote{For Llama 3 8B, we did our experiments using a Vast.ai instance which had an H100 as a GPU with a 100gb SSD; this took 10 minutes to complete for each experiment of 1000 trials ($\sim$40 minutes total).}. In all experiments, we recorded the final hands of the player and dealer, game outcomes (win, loss, tie), and all cards drawn during gameplay.

\subsection{Statistical Analysis}

To compare the distributions of outcomes between the control and LLM groups, we employed several statistical methods. 

\textbf{Kullback-Leibler (KL) Divergence}: Measures the difference between two probability distributions, quantifying the amount of information lost when using one distribution to approximate another. If the KL divergence between the two distributions is small, it indicates that the distributions are similar. We refer to the KL divergence between distributions $P$ and $Q$ as $D_{KL}(P \| Q)$ \cite{kullback1951information}.

\textbf{Chi-Squared Test}: Determines whether there is a statistically significant difference between the expected frequencies and the observed frequencies in one or more categories. We refer to the test statistic and corresponding p-value as $\chi^2$ and $p\text{-value}_\text{chi-squared}$ \cite{doi:10.1080/14786440009463897}. 

\textbf{K-Sample Anderson-Darling Test}: Assesses whether \(k\) samples of data come from the same probability distribution. We refer to the test statistic and corresponding p-value as $A^2$ and $p\text{-value}_\text{anderson-darling}$ \citep{engmann_adktest_2011}. 

VFA considers there to be a distribution shift between the LLM outcomes and the control outcomes exists if \( D_{KL}(P \| Q) \) is non-zero and both \( p\text{-value}_{\text{chi-squared}} \leq 0.05 \) and \( p\text{-value}_{\text{anderson-darling}} \leq 0.05 \).

\section{Results}

Our statistical analysis reveals significant distribution shifts in both card frequencies and final hand values between the LLM-controlled experiments and the theoretical baseline. \Cref{tab:card_frequencies} and \Cref{tab:final_hand_values} summarize these findings across different models and prompting strategies.


\begin{table*}[ht]
    \centering
    \caption{Statistical Test Results for Player and Dealer Card Frequencies}
    \label{tab:card_frequencies}
    \small
    \begin{tabular}{@{}llllllll@{}}
        \toprule
        \multirow{2}{*}{\textbf{Model}} & \multirow{2}{*}{\textbf{Shots}} & \multicolumn{2}{l}{\textbf{KL Divergence}} & \multicolumn{2}{l}{\textbf{Chi-Squared}} & \multicolumn{2}{l@{}}{\textbf{Anderson-Darling}} \\
        \cmidrule(lr){3-4} \cmidrule(lr){5-6} \cmidrule(l){7-8}
        & & \textbf{Dealer} & \textbf{Player} & \textbf{Dealer} & \textbf{Player} & \textbf{Dealer} & \textbf{Player} \\
        \midrule
        GPT-4 & Zero & 0.599 & 0.902 & 1,720$^{***}$ & 1,920$^{***}$ & 2.719$^{*}$ & 3.369$^{**}$\\
        GPT-4 & Few & 0.851 & 1.000 & 1,648$^{***}$ & 2,105$^{***}$ & 3.365$^{**}$ & 3.153$^{**}$\\
        Claude 3.5 & Zero & 3.128 & 3.287 & 5,548$^{***}$ & 4,337$^{***}$ & 4.566$^{**}$ & 4.591$^{**}$\\
        Claude 3.5 & Few & 7.735 & 5.330 & 11,394$^{***}$ & 5,872$^{***}$ & 7.933$^{***}$ & 6.158$^{**}$\\
        Llama 3 & Zero & 7.019 & 7.157 & 4,186$^{***}$ & 4,390$^{***}$ & 3.906$^{**}$ & 5.157$^{**}$\\
        Llama 3 & Few & 8.116 & 8.323 & 4,698$^{***}$ & 4,622$^{***}$ & 5.503$^{*}$ & 5.513$^{*}$\\
        \bottomrule
    \end{tabular}
    \begin{minipage}{.66\textwidth}
        \vspace{.2em}
        \footnotesize * $p \leq 0.05$, ** $p \leq 0.01$, *** $p \leq 0.001$.
    \end{minipage}
\end{table*}

\begin{table*}[ht]
    \centering
    \caption{Statistical Test Results for Player and Dealer Final Hand Values}
    \label{tab:final_hand_values}
    \small
    \begin{tabular}{@{}llllllll@{}}
        \toprule
        \multirow{2}{*}{\textbf{Model}} & \multirow{2}{*}{\textbf{Shots}} & \multicolumn{2}{l}{\textbf{KL Divergence}} & \multicolumn{2}{l}{\textbf{Chi-Squared}} & \multicolumn{2}{l@{}}{\textbf{Anderson-Darling}} \\
        \cmidrule(lr){3-4} \cmidrule(lr){5-6} \cmidrule(l){7-8}
        & & \textbf{Dealer} & \textbf{Player} & \textbf{Dealer} & \textbf{Player} & \textbf{Dealer} & \textbf{Player} \\
        \midrule
        GPT-4 & Zero & 0.253 & 0.512 & 345$^{***}$ & 307$^{***}$ & $-$0.236 & 0.783 \\
        GPT-4 & Few & 0.062 & 0.165 & 141$^{***}$ & 248$^{***}$ & $-$0.776 & 0.338 \\
        Claude 3.5 & Zero & 3.539 & 2.661 & 2,950$^{***}$ & 2,695$^{***}$ & 2.782$^{*}$ & 3.939$^{**}$ \\
        Claude 3.5 & Few & 7.246 & 4.147 & 5,284$^{***}$ & 3,390$^{***}$ & 12.826$^{***}$ & 6.296$^{**}$ \\
        Llama 3 & Zero & 1.316 & 1.318 & 6,967$^{***}$ & 1,661$^{***}$ & 2.337$^{*}$ & 1.856 \\
        Llama 3 & Few & 0.876 & 1.940 & 1,042$^{***}$ & 905$^{***}$ & 1.979$^{*}$ & 3.045$^{*}$ \\
        \bottomrule
    \end{tabular}
    \vspace{1ex}
    \begin{minipage}{.66\textwidth}
        \vspace{.2em}
        \footnotesize * $p \leq 0.05$, ** $p \leq 0.01$, *** $p \leq 0.001$.
    \end{minipage}
\end{table*}

\subsection{Card Frequencies}

\begin{figure*}
    \centering
    \includegraphics[width=.9\linewidth]{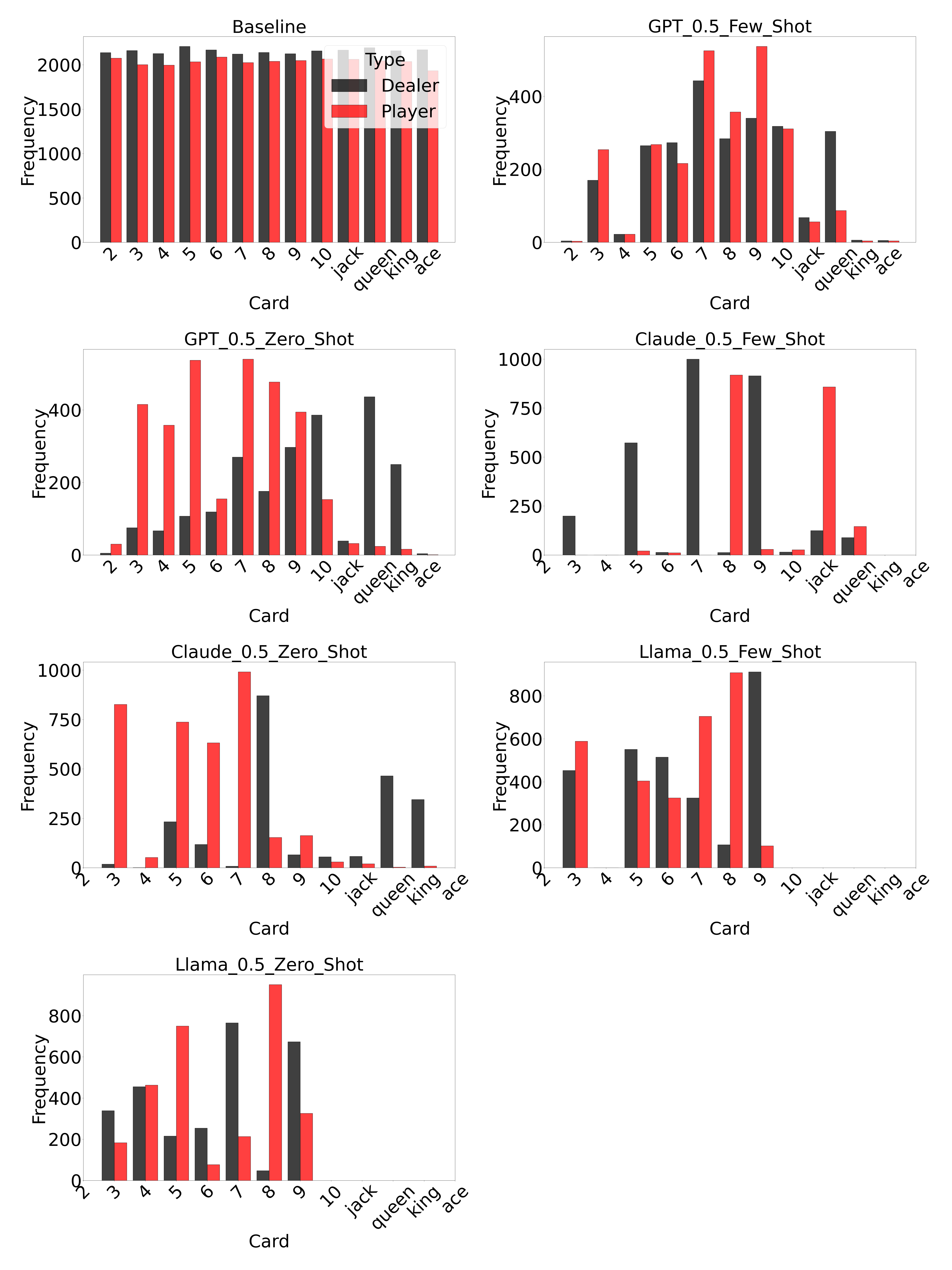}
    \caption{This figure shows the card draw frequencies across experiments. Compared to the baseline, all models exhibit significant deviations. Llama 3, for example, never outputs a face card, while Claude seems to choose specific cards far more frequently than others.}
    \label{fig:card_frequencies}
\end{figure*}

As shown in \Cref{fig:card_frequencies}, card draw frequencies vary considerably across models and prompting strategies. Kullback-Leibler (KL) divergence values are notably non-zero across all models and prompting strategies, indicating significant shifts in card draw frequencies (\Cref{tab:card_frequencies}). This observation is further corroborated by the chi-squared test, which consistently yield \(p\)-values below 0.001 for all models and settings, confirming statistically significant distributional shifts. The Anderson-Darling test results also show highly significant differences ($p\text{-values}\leq 0.01$) between observed and expected card frequencies in nearly all cases.

Interestingly, the magnitude of these shifts varies across models. GPT-4 demonstrates the lowest KL divergence values, ranging from 0.599 to 1.000, while Claude 3.5 and Llama 3 exhibit substantially higher values, ranging from 3.128 to 8.323. This suggests that while all models deviate from the theoretical baseline, some models, particularly GPT-4, may align more closely with expected card frequencies.

\subsection{Final Hand Values}

The distribution of final hand values also exhibits significant deviations from the baseline expectations. As shown in \Cref{tab:final_hand_values}, KL divergence values for final hand values are non-zero across all models and prompting strategies, indicating distributional shifts. Chi-squared test results consistently show \(p\)-values below 0.001, further confirming these shifts.

Anderson-Darling test results for final hand values show a pattern. While Claude 3.5 and Llama 3 show statistically significant deviations ($p\text{-values}\leq 0.05$) in most cases, GPT-4 yields non-significant results for both zero-shot and few-shot prompting. While GPT-4's final hand value distributions do deviate from the baseline, indicated by non-zero KL divergence and significant chi-squared results, these deviations may be less extreme compared to the other models.

It's worth noting that Claude 3.5 with few-shot prompting shows particularly high KL divergence and Anderson-Darling test statistic values for dealer final hand values, 7.246 and 12.826 respectively. This could indicate a more pronounced deviation in dealer behavior for this specific model and prompting strategy. A detailed visualization of hand value distributions is provided in \Cref{appendix:hand_value_plots}.

Our results collectively support the hypothesis that VFA can effectively highlight distribution shifts in LLM decision-making, within the context of a simplified blackjack game. The observed shifts vary in magnitude across different models and prompting strategies, suggesting potential differences in how LLM biases manifest.

\section{Discussion and Conclusion}

Our study uses blackjack as a proof-of-concept to test our framework's ability to detect deviations from theoretical baselines in LLM decision-making. Significant deviations in both card frequencies and final hand values were observed, with KL divergence, Chi-Squared, and Anderson-Darling test results supporting statistically significant shifts from expected distributions. These results are consistent with our hypothesis that the framework can effectively highlight anomalous LLM behavior, with \(p\)-values consistently below 0.001 across various models.


It is important to acknowledge that there are limitations to our work. The lack of human baseline data means our results on blackjack are based on comparisons to theoretical expectations of random card draws rather than actual human performance in the same tasks. Additionally, it is unclear whether the deviations observed in the blackjack environment will generalize to other, more complex environments. As such, the scope of our current findings is restricted to this specific, relatively narrow use case.

Our immediate next steps are to collect human baseline data for the blackjack task to more accurately assess LLM performance. This will allow us to directly compare LLM and human performance. In addition, we will expand our experiments to cover a broader range of environments to test the robustness and generality of our framework. Potential environments include financial decision-making scenarios, automated hiring systems, and strategic game simulations.

Observed performance differences between models such as GPT-4, Llama 3, and Claude 3.5 Sonnet suggest some hidden factor influencing model behavior. Future work will compare LLM variants, including instruction fine-tuned models, to further investigate the impact of training techniques on model behavior.

\section{Social Impact Statement}

Our framework for evaluating distribution shifts in model behavior has potential implications for developing safer AI systems. By detecting subtle biases or misalignments in LLM decision-making, this work could contribute to more robust evaluation practices for AI models in high-stakes domains. While our current results are limited to a simplified blackjack environment, the ability to identify distributional anomalies not apparent from individual outputs could help researchers better understand LLM limitations and failure modes. As language models become increasingly integrated into critical processes, approaches like ours could play a small but meaningful role in ensuring these systems behave as intended across various inputs and environments, potentially contributing to the development of more reliable and trustworthy AI technologies.

{\small
\bibliography{vfa}
\bibliographystyle{icml2024}
}

\newpage
\appendix
\onecolumn

\section{Simplified Blackjack Rules}
\label{appendix:blackjack_rules}

Our experiment uses a simplified version of blackjack with the following rules:\\[-1.8em]
\begin{enumerate}
    \item The game uses a standard 52-card deck, reshuffled before each hand.
    \item One player competes against the dealer.
    \item Card values:
    \begin{itemize}
        \item Face cards (Jack, Queen, King) = 10 points
        \item Aces = 1 or 11 points (whichever is more advantageous)
        \item All other cards = face value
    \end{itemize}
    \item Goal: Get as close to 21 points as possible without exceeding it.
    \item Game loop:
    \begin{itemize}
        \item Player and dealer each receive two cards; one dealer card is face-up
        \item Player can ``hit'' (take another card) or ``stand'' (keep current hand)
        \item Player loses immediately if their hand exceeds 21 (``bust'')
        \item After player stands, dealer hits until reaching 17 or higher
        \item Dealer must hit on a ``soft 17''(hand with an Ace counted as 11)
    \end{itemize}
    \item Win conditions:
    \begin{itemize}
        \item Player wins if dealer busts
        \item If neither busts, hand closest to 21 wins
        \item If both hands are the same, they tie
    \end{itemize}
    \item Simplifications
    \begin{itemize}
        \item No splitting pairs
        \item No doubling down
        \item No insurance bets
        \item No surrendering\\[-1.8em]
    \end{itemize}
\end{enumerate}
{\footnotesize Traditional Blackjack rules can be found at \url{https://bicyclecards.com/how-to-play/blackjack}}

\subsection{Basic Blackjack Playing Strategy}
\label{appendix:blackjack_strategy}

The player follows a basic strategy depending on the dealer's upcard and the player's hand value:
\begin{algorithm}
\caption{Player Strategy in Blackjack}
\begin{algorithmic}
    \REQUIRE dealer's upcard, player's hand value
    \WHILE{player's turn}
        \IF{dealer's upcard $\geq 7$ \AND player's hand value $<$ 17}
            \STATE Hit
            \STATE Update player's hand value
        \ELSIF{dealer's upcard $\leq 6$ \AND player's hand value $<$ 12}
            \STATE Hit
            \STATE Update player's hand value
        \ELSE
            \STATE Stand \COMMENT{Stop taking cards}
            \STATE End player's turn
        \ENDIF
    \ENDWHILE
\end{algorithmic}
\end{algorithm}

\newpage
\section{LLM Prompts}
\label{appendix:prompts}

Below are the prompts used for the simulations.

\subsection{Zero-Shot Prompt}
\begin{tcolorbox}[colback=gray!5!white, colframe=gray!75!black, title=Zero-Shot Prompt]
    You are a blackjack dealer at a casino. \\
    You will be responsible for drawing cards randomly for both the player and dealer. You will respond with the card that you draw as the dealer. Possible cards you can draw are [2, 3, 4, 5, 6, 7, 8, 9, 10, Jack, Queen, King, Ace]. \\
    You will be provided the game state. \\
    You will now randomly draw your cards from the deck to deal.
    
    \texttt{\{game\_state\}}

    Do not give any additional details. \\
    Your drawn card is 
\end{tcolorbox}

\subsection{Few-Shot Prompt}
\begin{tcolorbox}[colback=gray!5!white, colframe=gray!75!black, title=Few-Shot Prompt]
    You are a blackjack dealer at a casino. \\
    You will be responsible for drawing cards randomly for both the player and dealer. You will respond with the card that you draw as the dealer. Possible cards you can draw are [2, 3, 4, 5, 6, 7, 8, 9, 10, Jack, Queen, King, Ace]. \\
    You will be provided the game state. \\
    You will now randomly draw your cards from the deck to deal.

    Here are some examples of responses:

    Q: Your drawn card is \\
    A: Ace

    Q: Your drawn card is \\
    A: 4

    Q: Your drawn card is \\
    A: King

    \texttt{\{game\_state\}}

    Do not give any additional details. \\
    Your drawn card is 
\end{tcolorbox}

\vspace{1em}

\pagebreak

\newcommand{\scell}[3]{\begin{tabular}[#1]{@{}#2@{}}#3\end{tabular}}

\section{Game Statistics for each Model}
\label{appendix:game_statistics}
\begin{table}[h]
    \centering
    \caption{Summary of Model Results (1000 games per experiment)}
    \label{tab:summary_stats}
    \begin{tabular}{@{}lllll@{}}
        \toprule
        \bf Model & \scell{c}{l}{\bf Player\\\bf Win Rate} & \scell{c}{l}{\bf Dealer\\\bf Bust Rate} & \scell{c}{l}{\bf Average\\\bf Player Hand} & \scell{c}{l}{\bf Average\\\bf Dealer Hand} \\
        \midrule
        Expected & 0.425 & 0.244 & 18.75 & 19.72 \\
        GPT 4 (0.0, Zero-Shot) & 0.0 & 0.0 & 19.037 & 19.134 \\
        GPT 4 (0.0, Few-Shot) & 0.289 & 0.165 & 19.265 & 20.956 \\
        GPT 4 (0.5, Zero-Shot) & 0.359 & 0.102 & 19.755 & 18.867 \\
        GPT 4 (0.5, Few-Shot) & 0.399 & 0.205 & 19.46 & 19.017 \\
        Claude (0.0, Zero-Shot) & 1.0 & 0.0 & 21.0 & 18.0 \\
        Claude (0.0, Few-Shot) & 1.0 & 0.0 & 21.0 & 18.0 \\
        Claude (0.5, Zero-Shot) & 0.774 & 0.133 & 20.411 & 18.771 \\
        Claude (0.5, Few-Shot) & 0.744 & 0.001 & 21.175 & 18.094 \\
        Llama 3 (0.0, Zero-Shot) & 0.0 & 0.0 & 18.0 & 20.0 \\
        Llama 3 (0.0, Few-Shot) & 0.0 & 0.0 & 17.0 & 19.0 \\
        Llama 3 (0.5, Zero-Shot) & 0.216 & 0.156 & 18.854 & 18.543 \\
        Llama 3 (0.5, Few-Shot) & 0.347 & 0.347 & 18.646 & 17.237 \\
        \bottomrule
    \end{tabular}
    \begin{minipage}{0.695\textwidth}
        \vspace{.1em}
        \footnotesize \textit{Note: For Llama 3 at temperature 0.0 across zero and few-shot prompting, the dealer won every hand.}
    \end{minipage}
\end{table}

\section{Temperature Sampling Experiments}
\label{appendix:temperature}

\begin{table}[h]
    \centering
    \caption{Statistical Test Results for Player and Dealer Card Frequencies}
    \label{tab:temp0_card_frequencies}
    \begin{tabular}{@{}llllllllll@{}}
        \toprule
        \bf Model & \bf Temp. & \bf Shots & \multicolumn{2}{l}{\bf KL Divergence} & \multicolumn{2}{l}{\bf Chi-Squared Test} & \multicolumn{2}{l@{}}{\bf Anderson-Darling Test} \\
        \cmidrule(lr){4-5} \cmidrule(lr){6-7} \cmidrule(l){8-9}
        & & & \bf Dealer & \bf Player & \bf Dealer & \bf Player & \bf Dealer & \bf Player \\
        \midrule
        GPT 4 & 0.0 & Zero & 11.874 & 8.841 & 11869*** & 5788*** & 12.049*** & 6.347** \\
        GPT 4 & 0.0 & Few & 6.402 & 8.803 & 6976*** & 7596*** & 6.449** & 7.499*** \\
        Claude 3.5 & 0.0 & Zero & 13.077 & 10.696 & 14132*** & 5849*** & 13.624*** & 8.434*** \\
        Claude 3.5 & 0.0 & Few & 13.165 & 11.817 & 14665*** & 8271*** & 13.624*** & 10.605*** \\
        Llama 3 & 0.0 & Zero & 11.835 & 11.857 & 8872*** & 8376*** & 10.573*** & 10.605*** \\
        Llama 3 & 0.0 & Few & 11.921 & 11.830 & 9155*** & 8305*** & 10.573*** & 10.605*** \\
        \bottomrule
    \end{tabular}
    \begin{minipage}{0.84\textwidth}
    \vspace{.2em}
    \footnotesize \textit{Note: * $p \leq 0.05$, ** $p \leq 0.01$, *** $p \leq 0.001$.}
    \end{minipage}
\end{table}

\begin{table}[!hb]
    \centering
    \caption{Statistical Test Results for Player and Dealer Final Hand Values}
    \label{tab:temp0_final_hand_values}
    \begin{tabular}{@{}lllllllll@{}}
        \toprule
        \bf Model & \bf Temp. & \bf Shots & \multicolumn{2}{l}{\bf KL Divergence} & \multicolumn{2}{l}{\bf Chi-Squared Test} & \multicolumn{2}{l@{}}{\bf Anderson-Darling Test} \\
        \cmidrule(lr){4-5} \cmidrule(lr){6-7} \cmidrule(l){8-9}
        & & & \bf Dealer & \bf Player & \bf Dealer & \bf Player & \bf Dealer & \bf Player \\
        \midrule
        GPT 4 & 0.0 & Zero & 11.389 & 11.378 & 5329*** & 6372*** & 20.285*** & 16.466*** \\
        GPT 4 & 0.0 & Few & 2.688 & 9.401 & 2248*** & 4892*** & 5.445** & 13.034*** \\
        Claude 3.5 & 0.0 & Zero & 13.771 & 14.067 & 6144*** & 7829*** & 24.961*** & 21.324*** \\
        Claude 3.5 & 0.0 & Few & 13.771 & 14.067 & 6144*** & 7829*** & 24.961*** & 21.324*** \\
        Llama 3 & 0.0 & Zero & 14.065 & 13.679 & 7230*** & 6285*** & 24.961*** & 21.324*** \\
        Llama 3 & 0.0 & Few & 12.978 & 13.864 & 4275*** & 6947*** & 24.961*** & 21.323*** \\
        \bottomrule
    \end{tabular}
    \begin{minipage}{0.83\textwidth}
    \vspace{.2em}
    \footnotesize \textit{Note: The identical values for Claude 3.5 at temperature 0.0 across zero and few-shot prompting are due to both experiments consistently producing final hand values of 21 for the player and 18 for the dealer, resulting in the same values.}
    \end{minipage}
\end{table}

\section{Card Frequency Distributions Across Experiments}
\label{appendix:card_frequency_plots}

\begin{figure}[H]
    \centering
    \includegraphics[trim={5em 5.8em 10em 5em},clip,width=.8\linewidth]{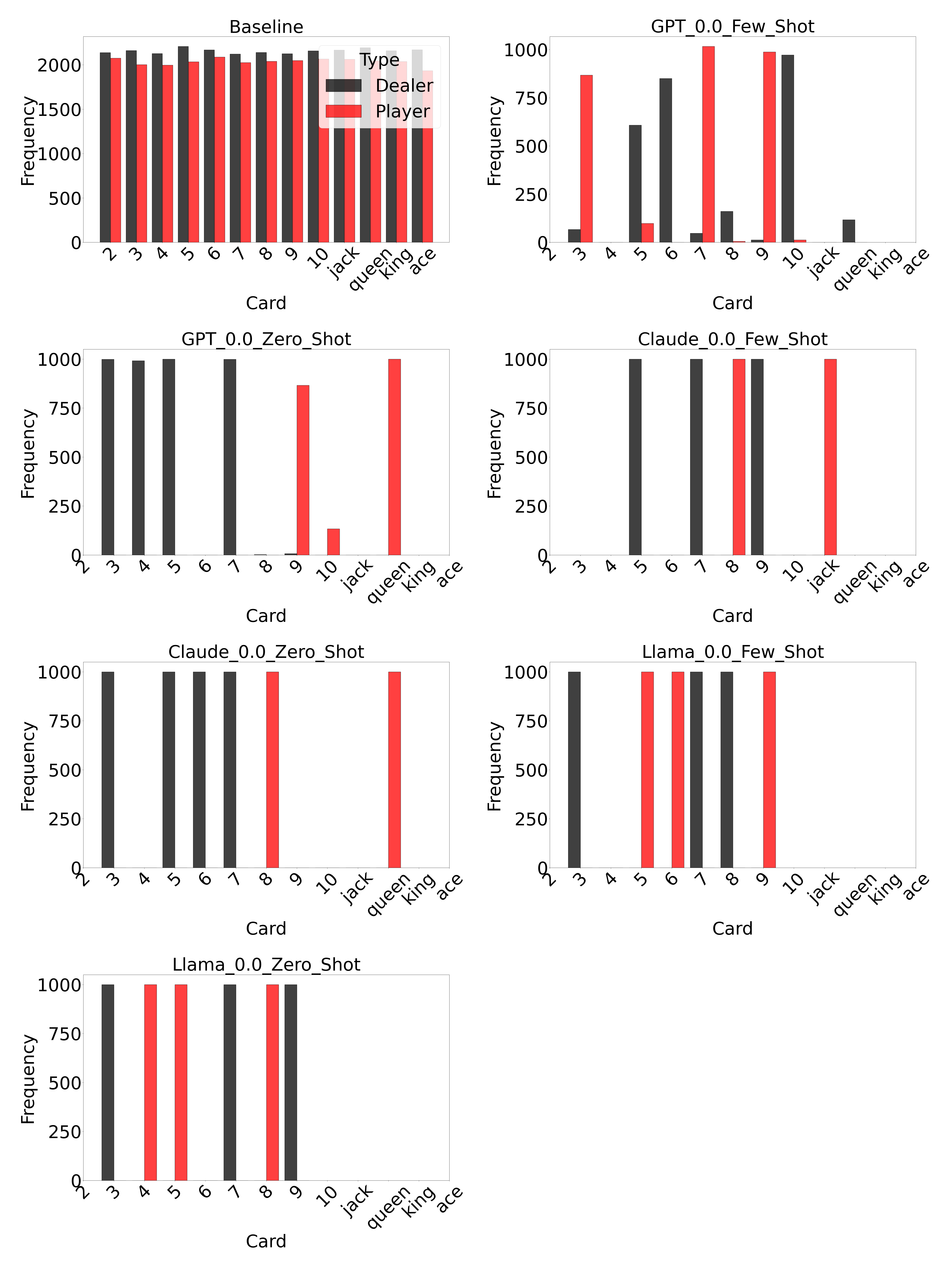}
    \caption{This figure compares the final hand value distribution for all experiments with temperature = 0.0. Compared to the baseline, the models tend to only draw cards between 1-9 and very rarely draw face cards.}
    \label{fig:card_frequency_plot_temp_0.0}
\end{figure}

\section{Final Hand Value Distributions Across Experiments}
\label{appendix:hand_value_plots}
\begin{figure}[H]
    \centering
    \includegraphics[trim={5em 5.8em 10em 5em},clip,width=.8\linewidth]{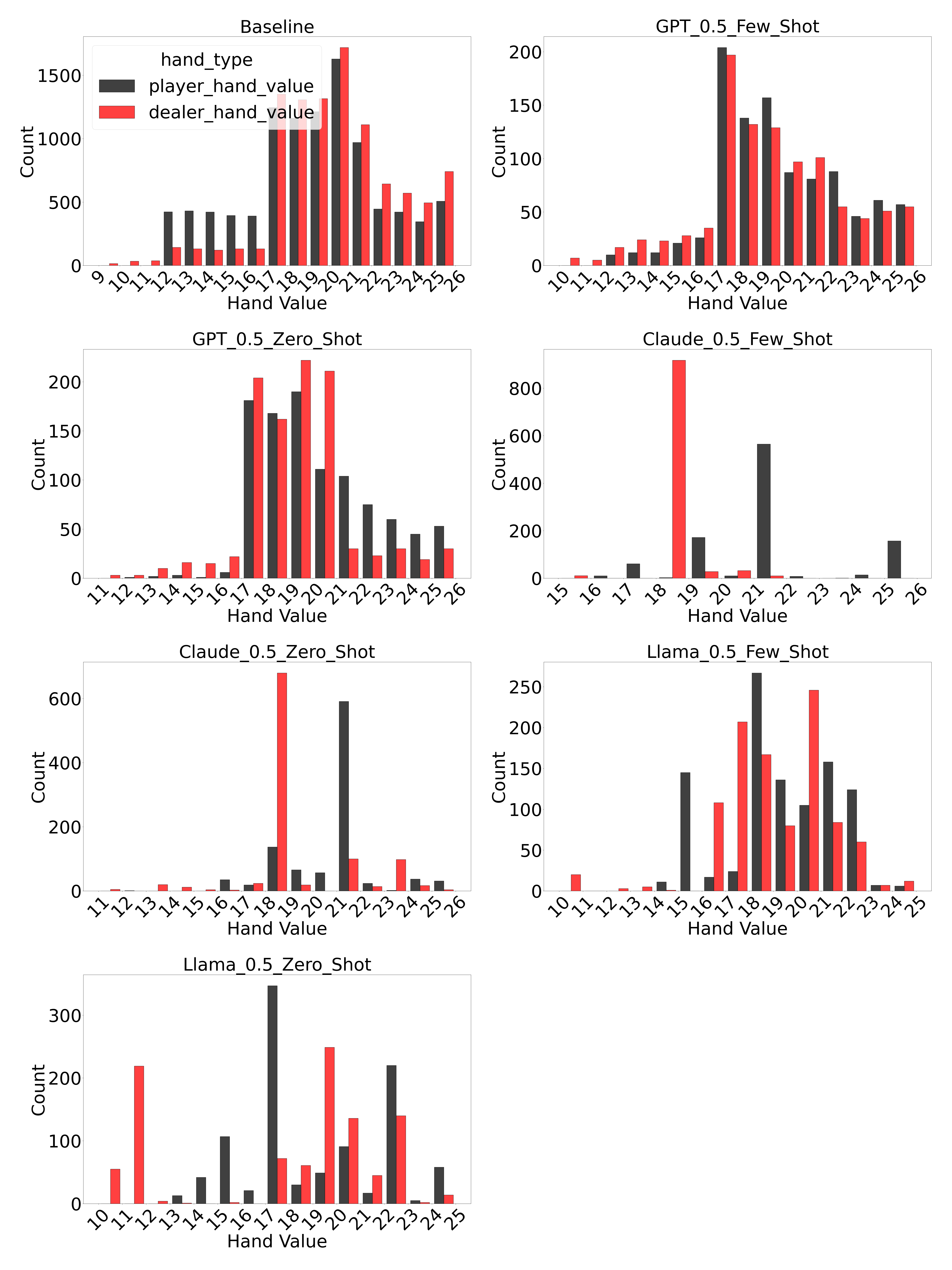}
    \caption{This figure compares the final hand value distribution for all experiments with temperature = 0.5. Compared to the baseline, the models demonstrate varying degrees of skewness. Claude's distribution is notably skewed, suggesting a preference for final hand values of 18 for the dealer and 21 for the dealer. In contrast, Llama and GPT's distributions appear to be closer to the expected distribution in a real blackjack game.}
    \label{fig:hand_value_plots_temp_0.5}
\end{figure}

\begin{figure}[H]
    \centering
    \includegraphics[trim={5em 5.8em 10em 5em},clip,width=.8\linewidth]{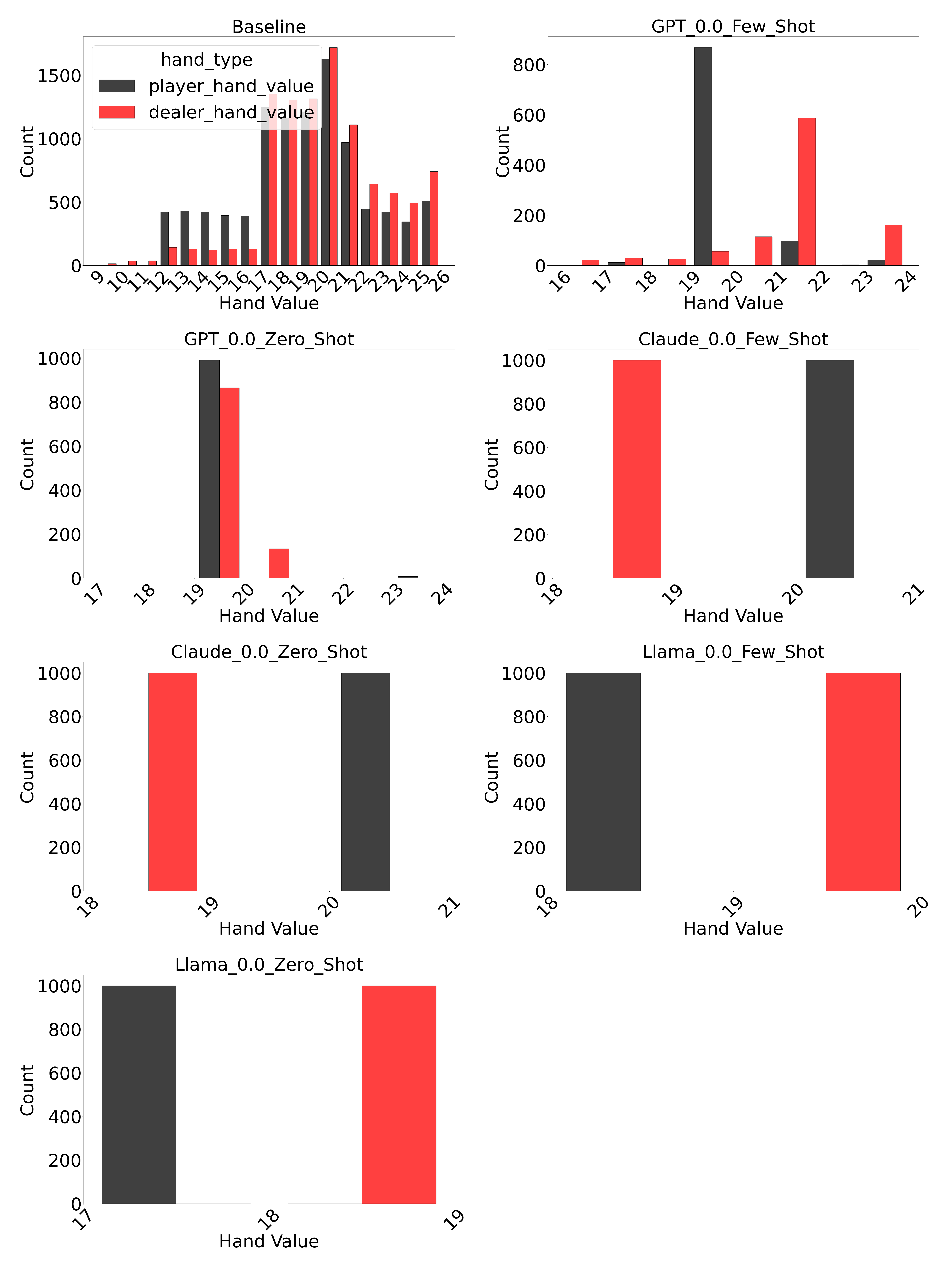}
    \caption{This figure compares the final hand value distribution for all experiments with temperature  = 0.0. Compared to the baseline, the models are very badly skewed. All models show tendencies to produce only one or two different kinds of hand value. Claude and Llama both only produce one value for the player and the dealer, ranging from 17-19 for the dealer and 19-21 for the player.}
    \label{fig:hand_value_plots_temp_0.0}
\end{figure}

\end{document}